\documentclass[letterpaper, 10 pt, journal, twoside]{IEEEtran}
\usepackage{amsmath,amsfonts}
\usepackage{algorithmic}
\usepackage{algorithm}
\usepackage{array}
\usepackage[caption=false,font=normalsize,labelfont=sf,textfont=sf]{subfig}
\usepackage{textcomp}
\usepackage{stfloats}
\usepackage{url}
\usepackage{verbatim}
\usepackage{graphicx}
\usepackage{cite}
\usepackage{amssymb}  
\usepackage{booktabs}  
\usepackage{arydshln}  
\usepackage{needspace}  
\usepackage{marvosym}  
\usepackage{hhline}
\usepackage{colortbl}
\usepackage{tabularx}
\usepackage{xcolor} 
\usepackage{multirow} 
\usepackage{hyperref}

\begin{document}

\title{ViTacGen: Robotic Pushing with Vision-to-Touch Generation}

\author{Zhiyuan Wu$^1$, Yijiong Lin$^2$, Yongqiang Zhao$^1$, Xuyang Zhang$^1$, Zhuo Chen$^1$, Nathan Lepora$^2$, Shan Luo$^1$
\thanks{This work was partially supported by the EPSRC project “ViTac: Visual-Tactile Synergy for Handling Flexible Materials” (EP/T033517/2).  (Corresponding author: Shan Luo.)}
\thanks{$^1$Zhiyuan Wu, Yongqiang Zhao, Xuyang Zhang, Zhuo Chen, and Shan Luo are with Department of Engineering, King's College London, Strand, London, WC2R 2LS, United Kingdom, \{zhiyuan.1.wu, yongqiang.zhao, xuyang.zhang, zhuo.7.chen, shan.luo\}@kcl.ac.uk. }
\thanks{$^2$Yijiong Lin and Nathan Lepora are with the Department of Engineering Mathematics and Bristol, Robotics Laboratory, University of Bristol, Bristol, BS8 1UB, United Kingdom, \{yijiong.lin, n.lepora\}@bristol.ac.uk. }
}


\maketitle

\begin{abstract}
Robotic pushing is a fundamental manipulation task that requires tactile feedback to capture subtle contact forces and dynamics between the end-effector and the object. However, real tactile sensors often face hardware limitations such as high costs and fragility, and deployment challenges involving calibration and variations between different sensors, while vision-only policies struggle with satisfactory performance. Inspired by humans' ability to infer tactile states from vision, we propose ViTacGen, a novel robot manipulation framework designed for visual robotic pushing with vision-to-touch generation in reinforcement learning to eliminate the reliance on high-resolution real tactile sensors, enabling effective zero-shot deployment on visual-only robotic systems. Specifically, ViTacGen consists of an encoder-decoder vision-to-touch generation network that generates contact depth images, a standardized tactile representation, directly from visual image sequence, followed by a reinforcement learning policy that fuses visual-tactile data with contrastive learning based on visual and generated tactile observations. We validate the effectiveness of our approach in both simulation and real world experiments, demonstrating its superior performance and achieving a success rate of up to 86\%. Code and data are available on \href{https://robot-perception-lab.github.io/vitacgen-website/}{https://robot-perception-lab.github.io/vitacgen-website/}.
\end{abstract}


\vspace{-0.8em}
\section{Introduction}
\IEEEPARstart{R}obotic pushing is a fundamental manipulation task that involves applying forces to move objects toward a specified target region \cite{deng2024coarse}. This task requires precise perception of the interactions between the robot and its environment during execution to enable accurate dynamic control \cite{yu2016more}. In recent years, data-driven reinforcement learning (RL) approaches relying primarily on visual input have been widely explored for robotic pushing tasks. Some vision-based methods attempt to estimate contact dynamics. For instance, a nearby camera observing a compliant gripper can infer applied forces from the gripper’s deformation \cite{zhu2025forcesforfree}. However, such approaches only provide indirect measurements of contact dynamics, which limits their accuracy and robustness. Therefore, researchers have introduced tactile sensing into robotic pushing tasks, leveraging its ability to provide detailed information about contact forces, surface properties, and interaction dynamics \cite{lloyd2021goal, yang2023sim}. Compared to vision, tactile sensing provides localized and precise feedback \cite{luo2017tactilesurvey3}, which is particularly important for capturing the fine-grained interactions required in robotic pushing tasks. By utilizing tactile modalities, these methods have demonstrated improved performance in capturing the complex relationships between the robot, object, and environment. 

However, deploying tactile sensing in real world faces significant challenges. First, high-quality tactile sensors like vision-based sensors \cite{yuan2017gelsight, donlon2018gelslim} are inaccessible in many scenarios due to hardware limitations and specialized setups, and even when available, they typically require precise calibration \cite{donlon2018gelslim} and remain prone to noise, wear, and physical damage during prolonged use \cite{yuan2017gelsight}. Also, there exist manufacturing inconsistencies across different tactile sensors, which could cause subtle variations and manifest as noises, impairing model performance when deployed to new tactile sensors \cite{lin2022tactilegym2}.
These limitations create a critical perception gap in tactile sensing deployment, and visual-only policies struggle to capture the subtle contact dynamics necessary for precise control, resulting in suboptimal performance in real-world pushing tasks. 

Humans, in contrast, have the remarkable ability to infer tactile states from visual information \cite{newell2005visual}. Inspired by this, we propose ViTacGen, a novel robot manipulation framework designed for visual robotic pushing with vision-to-touch generation in reinforcement learning to eliminate the reliance on high-resolution real tactile sensors.
Specifically, our method introduces two key components: an encoder-decoder vision-to-touch generation network \textbf{VT-Gen} that synthesizes tactile contact depth images \cite{lin2022tactilegym2}, a standardized tactile representation, directly from visual image sequence, and an RL network \textbf{VT-Con} that learns robust policies through feature fusion and contrastive learning based on visual and generated tactile observations. Our ViTacGen is trained in simulation by two steps: (1) training VT-Gen to generate tactile contact depth images on paired visual and tactile data collected from expert trajectories by a pre-trained RL network \cite{lygerakis2024m2curl} with visual and tactile observations, and (2) incorporating the frozen VT-Gen to train VT-Con based on visual and generated tactile observations. ViTacGen can perform zero-shot deployment on visual-only robotic systems, which eliminates the reliance on high-resolution real tactile sensors but still captures subtle dynamics and interactions between the end-effector and the object. Additionally, we solve the problem of manufacturing variations across different tactile sensors using contact depth maps as a standardized tactile representation. We conduct extensive experiments in both simulation and real world settings to validate the effectiveness of the proposed framework. The results demonstrate that ViTacGen achieves superior performance, achieving a success rate of up to 86\% and outperforming baseline methods. Code and data will be open-sourced once the paper is accepted. 

Our main contributions are summarized as follows:
\begin{itemize}
\item We propose ViTacGen, a novel robot manipulation framework designed for visual robotic pushing with vision-to-touch generation in RL to eliminate the reliance on high-resolution real tactile sensors, enabling effective zero-shot deployment on visual-only robotic systems.
\item We introduce an encoder-decoder vision-to-touch generation network VT-Gen that generates contact depth images as a standardized tactile representation directly from visual image sequence, with an RL network VT-Con that learns robust policies using feature fusion and contrastive learning on visual and generated tactile observations.
\item We demonstrate the effectiveness of our proposed methods in both simulation and real world, through extensive qualitative and quantitative experiments, supported by comprehensive ablation studies.
\end{itemize}

\vspace{-0.4em}
\section{Related Works} \label{sec:relatedwork}

\subsection{Visual-Tactile Representation Learning}
Humans possess an extraordinary ability to perceive physical properties such as hardness, roughness, and texture through tactile sensing, which plays a critical role in providing feedback for tasks like grasping and manipulation. Research has increasingly focused on leveraging tactile information for robotic applications \cite{luo2017tactilesurvey3}, with a significant body of work aiming to integrate tactile sensing into manipulation tasks, such as material classification and grasping \cite{cao2020stam, luo2017tactilesurvey3}. For example, tactile feedback has been widely utilized to improve grasping stability \cite{deng2020grasping2}, which places high demands on the quality and robustness of tactile representations.

There is a natural synergy between visual and tactile modalities: vision provides a global understanding of objects, while touch offers detailed, localized information \cite{lygerakis2024m2curl}. Combining these two modalities can yield richer and more comprehensive representations. With the rise of deep learning, recent research on visual-tactile fusion has shifted from simple data-level integration to joint representation learning. Luo \textit{et al.} \cite{luo2018vitac} pioneered the use of feature fusion for visual-tactile representations in cloth texture recognition. Subsequent studies \cite{cui2020vtfsa, chen2022vtt} have employed attention mechanisms to enhance the fusion of visual and tactile features, further improving performance in various tasks.

More recently, contrastive learning has emerged as a powerful paradigm, replacing traditional supervised approaches. Techniques such as Information Noise Contrastive Estimation (InfoNCE) \cite{oord2018infonce} and Momentum Contrast (MoCo) \cite{he2020moco} have been successfully applied to learn joint visual-tactile representations, as demonstrated in works like \cite{yang2022touchandgo} and \cite{wu2025convitac}. Additionally, in visual-tactile perception where both modalities are image-based, the structural similarity between visual and tactile images has been extensively leveraged to explore cross-modal generative modeling \cite{lee2019touching}. Generative models exploit this similarity to synthesize tactile data from visual input and infer tactile-related physical parameters such as friction \cite{cao2023vis2hap} and material properties \cite{yang2022touchandgo}, particularly enabling vision-to-touch generation that allows the study of tactile properties without requiring direct physical interaction.

\vspace{-0.8em}
\subsection{Robotic Pushing}
Robotic pushing refers to the task of manipulating objects by applying forces to push them into a specified target region \cite{deng2024coarse}.
It requires precise perception of the interactions between the robot and its environment during execution to enable accurate dynamic control \cite{yu2016more}. While vision-based policies can estimate explicit contact dynamics, they rely heavily on precise visual perception such as end-effector deformation for contact parameter estimation \cite{zhu2025forcesforfree}. To overcome these limitations, researchers have introduced tactile sensing into robotic pushing tasks, leveraging its ability to provide detailed information about contact forces, surface properties, and interaction dynamics \cite{lloyd2021goal, yang2023sim}. Lloyd \textit{et al.} \cite{lloyd2021goal} first employed high-resolution optical tactile sensors in reactive robotic pushing to capture interaction details between end-effector and object. Lin \textit{et al.} \cite{lin2022tactilegym2} and Yang \textit{et al.} \cite{yang2023sim} pioneered in leveraging sim-to-real strategies with tactile sensing to train RL policies for robotic pushing. Recently, Lin \textit{et al.} \cite{lin2023bitouch} investigated a set of bimanual pushing tasks tailored towards touch with RL. However, tactile sensing in real world applications depends on highly accurate vision-based tactile sensors \cite{lin2022tactilegym2}, which can be inaccessible in many scenarios due to hardware limitations and specialized setups, and even when available, they typically require precise calibration \cite{donlon2018gelslim} and remain prone to noise, wear, and physical damage during prolonged use \cite{yuan2017gelsight}. Also, there exist manufacturing inconsistencies across different tactile sensors, which could cause subtle variations and manifest as noises, impairing model performance when deployed to new tactile sensors \cite{lin2022tactilegym2}. In this paper, we overcome these gaps by using generated contact depth maps from vision as a standardized tactile representation. 

\begin{figure*} [t!]
	\centering
    \vspace{0.6em}
	\includegraphics[width=0.7 \textwidth]{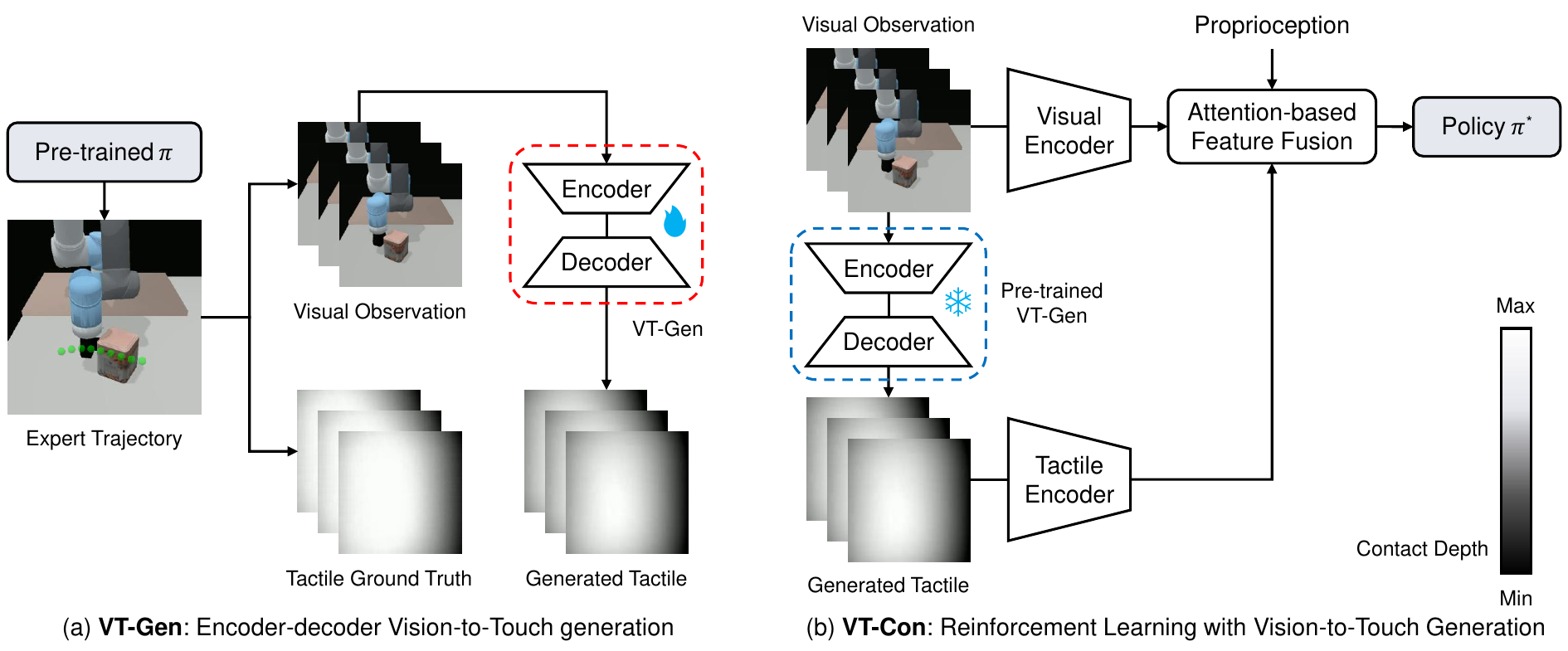}
    \vspace{-1.2em}
	\caption{
        The workflow of our proposed ViTacGen comprises two components: a VT-Gen for vision-to-touch generation, and a VT-Con for reinforcement learning on visual and generated tactile contact depth images with contrastive learning, and is trained in two stages: (a) In simulation, we collect paired visual and tactile data from expert trajectories by a pre-trained RL network \cite{lygerakis2024m2curl} with visual and tactile observations, and use them to train a vision-to-touch generation network VT-Gen so as to generate contact depth images as a standardized tactile representation from visual input, and (b) we incorporate the frozen VT-Gen to train a goal-conditioned RL policy VT-Con based on proprioception \textit{i.e.}, robot's TCP coordinates, visual, and generated tactile observations. This architecture enables ViTacGen to perform zero-shot deployment in real world using only visual information during inference. Please note that we employ contact depth \cite{church2022tactilegym} to represent our tactile images following \cite{lin2022tactilegym2} (see Sec. \ref{sec:tactile_gym_simulator} for details). 
    }\label{fig:pipeline}
    \vspace{-1.2em}
\end{figure*}

\vspace{-0.4em}
\section{Preliminaries}
\subsection{Reinforcement Learning}
Reinforcement learning (RL) has been widely applied in robot manipulation tasks, enabling robots to learn complex behaviors through trial and error. Among various RL algorithms, the Soft Actor-Critic (SAC) algorithm \cite{haarnoja2018sac} is a popular choice due to its ability to balance exploration and exploitation effectively. SAC is an off-policy actor-critic framework that incorporates entropy regularization into its objective function. This regularization encourages the policy to explore diverse actions, improving robustness and stability during training. Being off-policy, SAC leverages a replay buffer to learn from past experiences, which enhances sample efficiency \cite{lygerakis2024m2curl}. The objective function $J$ of SAC under policy $\pi$ is defined as:  
\begin{equation}  
J(\pi) = \mathbb{E}_{(s_t, a_t) \sim \rho_\pi} \left[ \sum_t \gamma^t \left( R(s_t, a_t) + \alpha \mathcal{H}(\pi(\cdot | s_t)) \right) \right],  
\end{equation}  
where $\mathbb{E}$ represents the expectation, $s_t$ represents the environmental state at time step $t$, including the temporal window of visual and generated tactile contact depth observation sequences $\mathcal{V}$ and $\mathcal{C}$, from time steps $t-2$ to $t$ (3 frames), as well as proprioception, \textit{i.e.}, robot's tool center point (TCP) coordinates, $a_t$ represents the robot's action, $\rho_\pi$ represents the state-action distribution, $\gamma$ represents the discount factor, $R$ represents the reward function, $\alpha$ represents the temperature parameter controlling the weight of the entropy term $\mathcal{H}$, and $\mathcal{H}(\pi(\cdot | s_t))$ denotes the entropy of the policy $\pi$ at state $s_t$:  
\begin{equation}  
\mathcal{H}(\pi(\cdot | s_t)) = -\mathbb{E}_{a_t \sim \pi(\cdot | s_t)} \left[ \log \pi(a_t | s_t) \right]. 
\end{equation} 
The reward function $R(s_t, a_t) = -d_{\text{goal}} - d_{\text{TCP}}$ is designed as a dense reward that minimizes both the object-to-goal distance $d_{\text{goal}}$ and the TCP-to-object distance $d_{\text{TCP}}$.

\subsection{Tactile Gym Simulator}
\label{sec:tactile_gym_simulator}
We use Tactile Gym 2 \cite{lin2022tactilegym2} as our simulation platform, which provides efficient visual-tactile robot learning and manipulation capabilities through high-resolution optical tactile sensors. In Tactile Gym 2, the robotic pushing task is defined as moving the object along a desired 2D trajectory through a series of goal points. The observation space includes visual sensing, tactile sensing, and proprioception, \textit{i.e.}, robot's TCP coordinates. The principle of its tactile simulator \cite{church2022tactilegym} is to obtain contact depth images as a standardized tactile representation across different tactile sensors in our simulation environment. Based on PyBullet \cite{coumans2016pybullet}, it utilizes synthetic depth camera rendering to capture contact depth images within a virtual optical tactile sensor. When contacting with an object, the system computes the difference between the current depth image and a reference depth image taken when the sensor is not in contact to produce a penetration depth map that generalizes to arbitrary sensor shapes. To achieve computational efficiency for large-scale data generation, the simulator approximates the soft tip of the tactile sensor using rigid body physics, limiting contact stiffness and damping parameters during collision detection to allow objects to penetrate the simulated sensor tip in a manner that approximates the deformation of a real tip. 

\vspace{-0.4em}
\section{Methodology}
\subsection{Overview}
As illustrated in Fig. \ref{fig:pipeline}, our proposed ViTacGen framework comprises two key components: \textbf{VT-Gen} for encoder-decoder vision-to-touch generation and \textbf{VT-Con} to learn robust policies through feature fusion and contrastive learning based on visual and generated tactile observations. The entire pipeline is correspondingly structured into two sequential phases to ensure effective training (simulation) and deployment (real world). Firstly, during training in simulation, we collect paired visual and tactile data from expert trajectories by a pre-trained RL network \cite{lygerakis2024m2curl} with visual, tactile, and proprioception observations, and use them to train the vision-to-touch generation network VT-Gen to generate contact depth images from a visual image sequence. Secondly, we incorporate the frozen VT-Gen to train the RL policy VT-Con based on visual and generated tactile observations. This pipeline enables zero-shot transfer from simulated training to real world inference on visual-only robotic systems without real tactile sensors. The subsequent sections offer an in-depth analysis of both VT-Gen and VT-Con's architectures.

\subsection{VTGen: Encoder-decoder Vision-to-touch Generation} \label{sec:vt-gen}
Humans have the remarkable ability to infer tactile states from visual information \cite{newell2005visual}. Inspired by this, we propose a VT-Gen for encoder-decoder vision-to-touch generation to infer the tactile sensation of the contact region directly from a image sequence of visual input. As depicted in Fig. \ref{fig:pipeline} (a), given a visual image sequence $\mathcal{V} = \{ \boldsymbol{v}_1, \dots, \boldsymbol{v}_{N} \}$, it is processed by VT-Gen to generate a current tactile contact depth image $\boldsymbol{c}^{gen}$, which is then repeated $N$ times to form a generated tactile contact depth sequence $\mathcal{C}$ with $N$ frames to align with the length of the visual sequence, serving as the tactile observation for VT-Con. 

Specifically, we first concatenate $\mathcal{V}$ along the channel dimension and extract features with coarse-to-refine encoding, so as to capture hierarchical and multi-scale visual features at multi scales \cite{wu2024s3mnet}. We pass $\mathcal{V}$ through a coarse encoder $\mathcal{E}_{coarse}$, which extracts the initial visual feature map $\boldsymbol{f}^v_{coarse}$. This is followed by a cross-modal attention operation $\mathcal{A}_{cm}$ \cite{dosovitskiy2020vit}, formulated as:
\begin{equation} \label{eq:attention}
    \boldsymbol{f}^v_{cm} = \boldsymbol{C}[\mathcal{A}^{cm}_1(\boldsymbol{f}^v_{coarse}, \boldsymbol{p}), \dots, \mathcal{A}^{cm}_h(\boldsymbol{f}^v_{coarse}, \boldsymbol{p})]\boldsymbol{w}_0, 
\end{equation}
where $\boldsymbol{p}$ represents a learnable positional embedding \cite{wu2025cdi3d}, $\boldsymbol{f}^v_{cm}$ refers to the processed visual feature map, $h$ is the number of attention heads that empirically set to $8$, and $\boldsymbol{w}_0$ represents the weight matrix used for output. The cross-modal attention operation $\mathcal{A}^{cm}_i$ is defined as:
\begin{equation} \label{eq:cross_modal}
    \mathcal{A}_{cm}(\boldsymbol{x}, \boldsymbol{y}) = softmax(\frac{\boldsymbol{q}\boldsymbol{k}^T}{\sqrt{d}}) \cdot \boldsymbol{v}, 
\end{equation}
with 
\begin{equation}
    \boldsymbol{q} = \boldsymbol{w}_q \cdot \boldsymbol{x}, \quad \boldsymbol{k} = \boldsymbol{w}_k \cdot \boldsymbol{y}, \quad \boldsymbol{v} = \boldsymbol{w}_v \cdot \boldsymbol{y}, 
\end{equation}
where $\boldsymbol{w}$ are learnable projection matrices \cite{dosovitskiy2020vit}. $\boldsymbol{f}^v_{cm}$ is then processed by a refine encoder $\mathcal{E}_{refine}$, to produce a deeper and more refined visual feature map $\boldsymbol{f}^v_{refine}$. Next, $\boldsymbol{f}^v_{refine}$ is processed through a series of identical residual blocks \cite{he2016resnet}, which enhance feature representation while preserving spatial information. Finally, a hierarchical decoder, composed of transposed convolutional layers and upsampling operations, generates the final contact depth prediction $\boldsymbol{c}^{gen}$. 

\subsection{VT-Con: RL with Visual-Tactile Contrastive Learning}  

\subsubsection{Feature Extraction}
\label{sec:feature_extraction}
As shown in Fig. \ref{fig:pipeline} (b), we consider the visual sequence $\mathcal{V} = \{ \boldsymbol{v}_1, \dots, \boldsymbol{v}_{N} \}$ for visual observation, and the generated tactile sequence $\mathcal{C} = \{ \boldsymbol{c}_1, \dots, \boldsymbol{c}_{N} \}$ for tactile observation, where $N$ denotes the number of frames in the visual and tactile sequences, respectively, empirically set to 3 for capturing the velocity and acceleration information. Each element $\boldsymbol{v}_i$ or $\boldsymbol{c}_i$ is represented as a tensor resized to the same dimension $\mathbb{R}^{H \times W \times C}$, where $H$, $W$, and $C$ correspond to the height, width, and channel count of the frames. To extract features from the visual and tactile inputs, we first concatenate $\mathcal{V}$ and $\mathcal{C}$ respectively along their channel dimensions. Then, we employ two structurally identical Convolutional Neural Networks (CNNs) $\mathcal{E}^v$ and $\mathcal{E}^c$ as encoders for the visual and tactile modalities, resulting in a visual feature map $\boldsymbol{f}^v$ and a tactile feature map $\boldsymbol{f}^c$.

\begin{figure} [t!]
	\centering
    \vspace{0.6em}
	\includegraphics[width=0.35 \textwidth]{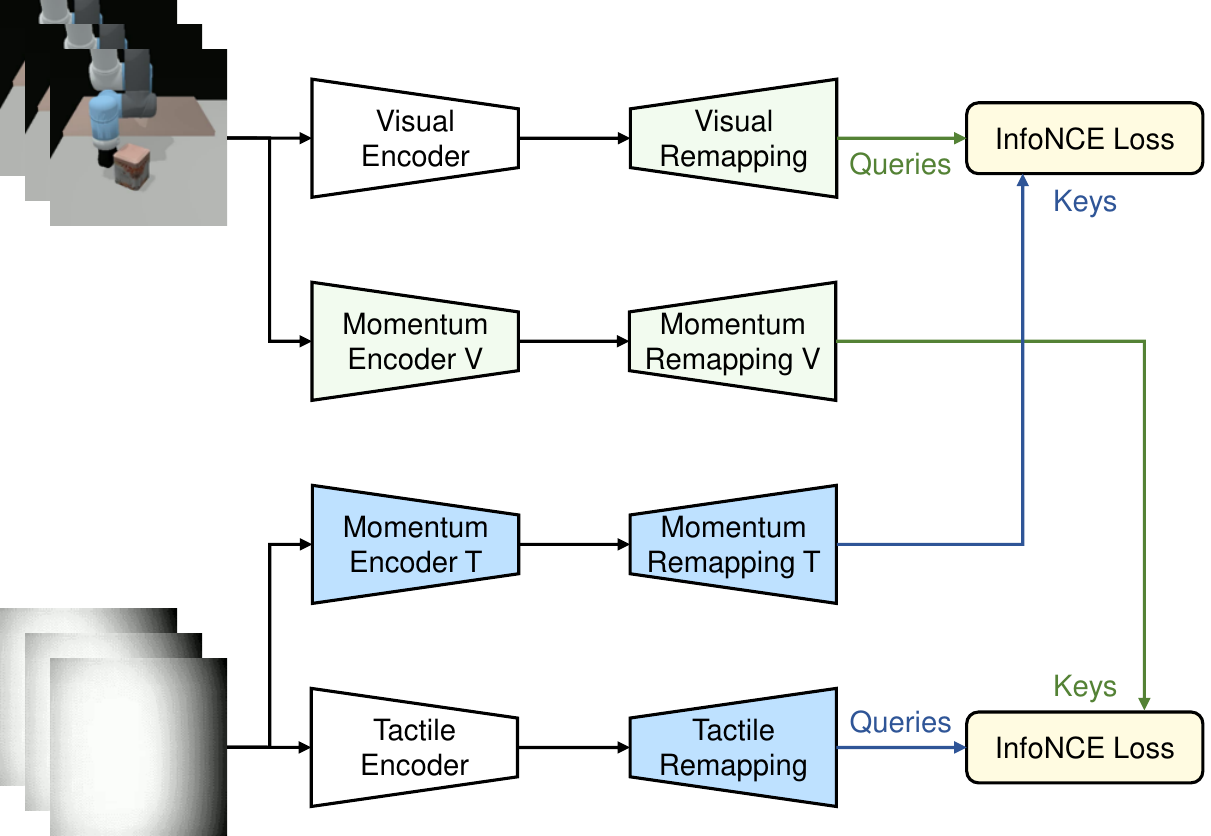}
    \vspace{-0.8em}
	\caption{
        To better fuse the encoded visual and tactile features, we optimize the visual and tactile encoders in VT-Con (Fig. \ref{fig:pipeline} (b)) with MoCo \cite{he2020moco} for contrastive learning between visual and tactile modalities. We utilize two momentum-based encoders to obtain momentum visual and tactile features, which are then aligned with the original features of the other modality through the InfoNCE loss \cite{oord2018infonce}.
    }\label{fig:cl_pipeline}
    \vspace{-1.2em}
\end{figure}

\subsubsection{Contrastive Learning between Vision and Touch}  
In order to more efficiently fuse the encoded visual and tactile features $\boldsymbol{f}^v$ and $\boldsymbol{f}^c$, we optimize $\mathcal{E}^v$ and $\mathcal{E}^c$ with Momentum Contrast (MoCo) \cite{he2020moco} for contrastive learning between vision and touch, inspired by \cite{lygerakis2024m2curl}. As shown in Fig. \ref{fig:cl_pipeline}, we utilize two momentum-based encoders $\mathcal{M}^v$ and $\mathcal{M}^c$ to obtain momentum visual and tactile features $\boldsymbol{m}^v$ and $\boldsymbol{m}^c$ from $\mathcal{V}$ and $\mathcal{C}$, respectively. These momentum encoders are structurally identical to the original encoders $\mathcal{E}^v$ and $\mathcal{E}^c$ but are updated using a momentum-based approach, formulated as:  
\begin{equation}
    \mathcal{M}^v \leftarrow \eta \mathcal{M}^v + (1 - \eta) \mathcal{E}^v, \quad \mathcal{M}^c \leftarrow \eta \mathcal{M}^c + (1 - \eta) \mathcal{E}^c,  
\end{equation}
where $\eta \in [0, 1)$ is the momentum coefficient that controls the update rate. We compute the contrastive loss between the visual and tactile modalities using the InfoNCE loss \cite{oord2018infonce}, aligning $\boldsymbol{f}^v$ with $\boldsymbol{m}^c$ and $\boldsymbol{f}^c$ with $\boldsymbol{m}^v$. (see Sec. \ref{sec:loss} for details). This contrastive learning module is integrated into VT-Con's RL process by incorporating the contrastive loss into the actor's training. Specifically, after each time the RL updates, the contrastive loss is computed, and a back-propagation is performed to optimize the encoders $\mathcal{E}^v$ and $\mathcal{E}^c$.

\subsubsection{Attention-based Visual-Tactile Feature Fusion}
We introduce an attention-based visual-tactile feature fusion operation \cite{dosovitskiy2020vit}, similar to Eq. \ref{eq:attention}, to integrate $\boldsymbol{f}^v$ and $\boldsymbol{f}^c$, which can be formulated as follows:  
\begin{equation}
    \boldsymbol{f}^{fuse} = \boldsymbol{C}[\mathcal{A}^{cm}_1(\boldsymbol{f}^v, \boldsymbol{f}^c), \dots, \mathcal{A}^{cm}_h(\boldsymbol{f}^v, \boldsymbol{f}^c)]\boldsymbol{w}_0, 
\end{equation}
where $\boldsymbol{f}^{fuse}$ refers to the fused feature map. Along with the fused feature map $\boldsymbol{f}^{fuse}$, the flattened and MLP-processed proprioception, \textit{i.e.}, robot's TCP coordinates, are concatenated to form the complete observation vector for robot reinforcement learning and can be compatible with various RL algorithms, such as SAC and PPO. 

\subsection{Loss Function}
\label{sec:loss}
\subsubsection{Loss Function for VT-Gen}  
In VT-Gen, we utilize the VGG loss \cite{simonyan2014vggloss} to supervise the training of our vision-to-touch generation. The VGG loss is designed to evaluate the perceptual similarity between the generated contact depth image $\boldsymbol{c}^{gen}$ and the ground truth contact depth image $\boldsymbol{c}^{gt}$ by comparing their high-level feature representations extracted from a pre-trained VGG network. Specifically, the loss $\mathcal{L}^{vgg}$ is defined as:  
\begin{equation}  
    \mathcal{L}^{vgg} = \| \phi(\boldsymbol{c}^{gen}) - \phi(\boldsymbol{c}^{gt}) \|^2,  
\end{equation}  
where $\phi$ denotes the feature extraction process using the encoder of a pre-trained VGG network \cite{simonyan2014vggloss}, $\boldsymbol{c}^{gt}$ represents the ground truth contact depth image, and $\| \cdot \|^2$ corresponds to the mean squared error.

\subsubsection{Loss Function for VT-Con}
In VT-Con, we align visual and tactile modalities using InfoNCE loss \cite{oord2018infonce} as the contrastive loss $\mathcal{L}^{c}$, which consists of a vision-to-touch loss $\mathcal{L}^{vt}$ and a touch-to-vision loss $\mathcal{L}^{tv}$, defined as:
\begin{equation}
    \mathcal{L}^{vt} = - \frac{1}{B} \sum_{i=1}^B \log \frac{ \exp ( \frac{ \boldsymbol{f}^v_i \cdot \boldsymbol{m}^c_i }{\tau} ) }{\sum_{\substack{j=1 \\ j \neq i}}^{B}   \exp ( \frac{ \boldsymbol{f}^v_i \cdot \boldsymbol{m}^c_j } {\tau} ) },
\end{equation}
and
\begin{equation}
    \mathcal{L}^{tv} = - \frac{1}{B} \sum_{i=1}^B \log \frac{ \exp ( \frac{ \boldsymbol{f}^c_i \cdot \boldsymbol{m}^v_i }{\tau} ) }{\sum_{\substack{j=1 \\ j \neq i}}^{B} \exp ( \frac{ \boldsymbol{f}^c_i \cdot \boldsymbol{m}^v_j } {\tau} ) },
\end{equation}
where $B$ denotes the number of samples (or batch size in implementation), and $\tau$ is the temperature parameter, empirically set to $0.1$. The overall contrastive loss $\mathcal{L}^{c}$ is then expressed as:
\begin{equation}
    \mathcal{L}^{con} = \mathcal{L}^{vt} + \mathcal{L}^{tv}. 
\end{equation}
This loss enables our VT-Con to optimize $\mathcal{E}^v$ and $\mathcal{E}^c$ and enhance the multi-modal understanding capability during the RL training, fusing visual and tactile features more efficiently. 

\vspace{-0.4em}
\section{Experiments}

\begin{figure} [t!]
	\centering
    \vspace{0.6em}
	\includegraphics[width=0.25 \textwidth]{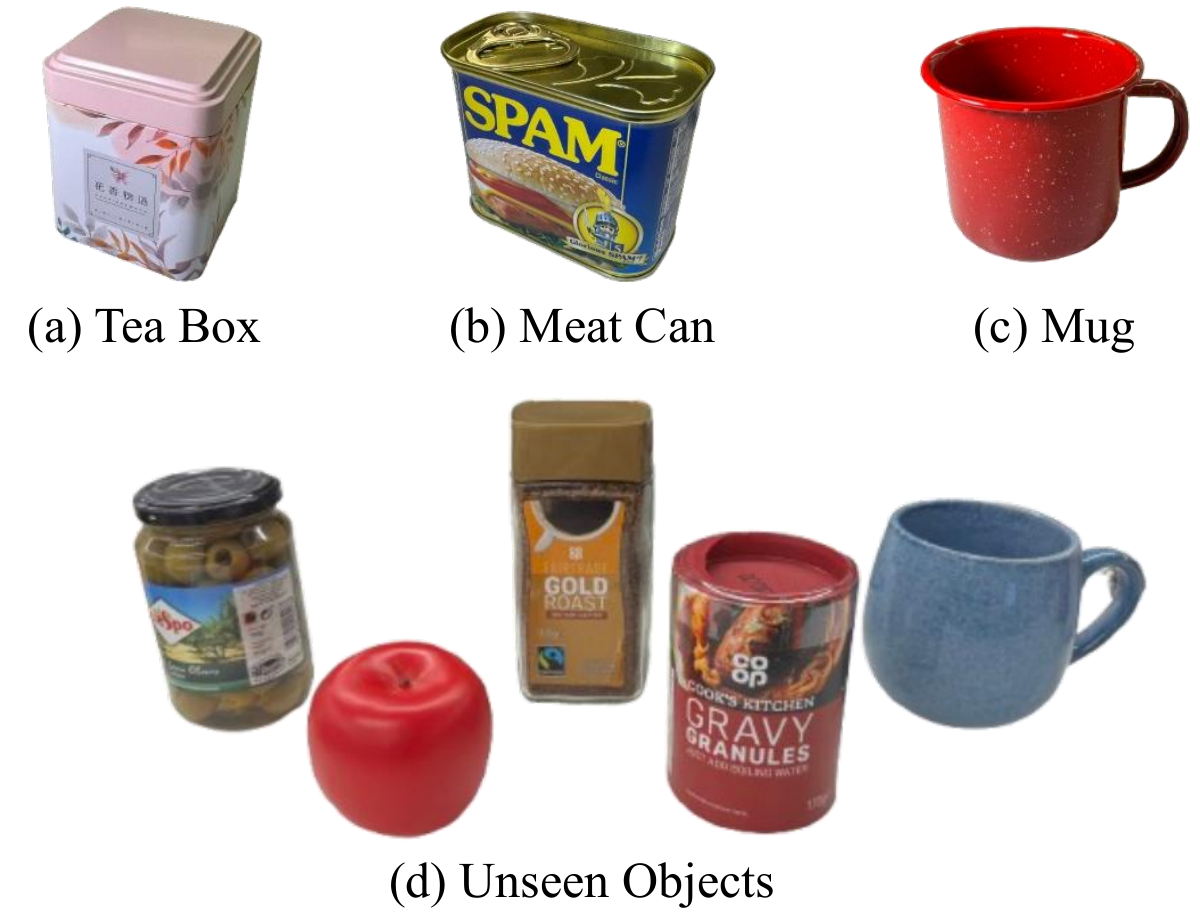}
    \vspace{-1.2em}
	\caption{
        Real world objects used in our experiments: (a) tea box, (b) meat can, (c) mug, and (d) unseen objects for zero-shot test including i) olive jar, ii) apple, iii) coffee can, iv) soup can, and v) ceramic cup. 
    }\label{fig:object}
    \vspace{-1.2em}
\end{figure}

\subsection{Experimental Setup}  
To validate the effectiveness of ViTacGen, we conducted comprehensive evaluations in both simulated and real world environments. Our simulation experiments were performed using Tactile Gym 2 \cite{lin2022tactilegym2} (see Sec. \ref{sec:tactile_gym_simulator} for details). We selected three representative objects: a tea box, a potted meat box, and a mug from the YCB dataset \cite{calli2015ycb}, whose digitally sampled meshes were incorporated into our simulation environment. We also selected five real-world unseen objects for testing. The objects are shown in Fig. \ref{fig:object}. We employ a UR5e robot arm equipped with a pushing end-effector to manipulate objects with a control frequency of 500Hz, with each manipulation episode limited to 350 timesteps within a constrained workspace of 800 $\times$ 600 mm in xy-plane at a fixed height of 2 cm above the table surface. We employ an external Intel RealSense D435 RGB camera with a field of view (FOV) of 42°, mounted 1 meter away from the robot base center with a 30° downward viewing angle, and the camera parameters are kept consistent in the corresponding simulation training. Our simulation experiments adhered to the standardized configurations of Tactile Gym 2 with a maximum of 350 time-steps per episode, and we implemented object mass, surface friction coefficient, and visual domain randomization including camera view, lighting, background, and color variations \cite{tobin2017domain} to improve our model's adaptability and robustness. The pushing trajectories are randomly generated using OpenSimplex noise.

For our VT-Gen, we collect visual and tactile data from 1,000 manipulation sequences and train our model for 200 epochs with a batch size of 64, splitting the data in a ratio of 7:2:1 for training, validation, and testing. For VT-Con, we employ the Soft Actor-Critic (SAC) \cite{haarnoja2018sac} algorithm from Stable Baselines 3 \cite{raffin2021sb3} as our RL backbone, with a max episode length of 350. We train our model for 1,000,000 time-steps with a batch size of 64 and a buffer size of 20,000 for experience replay. Both visual and tactile images are resized to 128 × 128 pixels. The success termination criterion is defined as the distance between the object center and the goal center being less than 2.5 cm. For optimizer we employ Adam \cite{kingma2014adam}, with a learning rate of 1e-4, and an epsilon value of 1e-8.

\subsection{Evaluation Metrics and Baselines}  
To evaluate the performance of our ViTacGen in both simulation and real world scenarios, we employ several key metrics. In the simulation environment, we assess cumulative rewards and episode length to evaluate the effectiveness of our learning algorithm. For both simulation and real-world experiments, we measure the final distance error, which quantifies the accuracy of the robot's manipulation tasks. Additionally, we calculate the success rates based on this distance error, with a threshold of less than 2.5 cm. For baseline comparison, in simulation we compare our method with visual-only policy, tactile-only policy, and visual \& tactile policy in \cite{lin2022tactilegym2}. We evaluate the performance of our ViTacGen with 1) visual-only input and 2) visual \& actual tactile input. The latter uses attention and contrastive learning, but without vision-to-touch generation. In real world, we compare our method with visual-only policy as we deploy our ViTacGen in a visual-only robotic system. 

\begin{table}[t!]  
    \centering  
    \caption{
    Quantitative results for vision-to-touch generation in RL setting. We report PSNR, SSIM \cite{wang2004ssim}, and LPIPS \cite{zhang2018lpips}.
    }  
    \label{tab:vtgnet}  
    \vspace{-1.2em}
    \begin{tabular}{c c c c}
        \toprule  
        Object Type & PSNR $\uparrow$ & SSIM $\uparrow$ & LPIPS $\downarrow$ \\
        \hline  
        Tea Box & 30.75 & 0.9482 & 0.0101 \\
        Meat Box & 20.50 & 0.8657 & 0.0327 \\
        Mug & 20.25 & 0.8222 & 0.0417 \\
        \bottomrule
    \end{tabular}  
    \vspace{-1.2em}
\end{table}  

\subsection{Effectiveness of Vision-to-Touch Generation}
\label{sec:exp_vtgnet}
To begin with, we quantitatively evaluate the effectiveness of our VT-Gen using standard metrics including PSNR, SSIM \cite{wang2004ssim}, and LPIPS \cite{zhang2018lpips}. As shown in Tab. \ref{tab:vtgnet}, VT-Gen achieves great performance on generating tactile contact depth map from visual image sequence, as evidenced by the tactile contact depth maps shown in Fig. \ref{fig:viz_sim} (ii) and (iii). Its robust performance can be attributed to its coarse-to-refine cross-modal based encoders, which capture hierarchical and multi-scale visual features. The high generation quality achieved by ViTacGen with visual-only input also demonstrates its effectiveness in maintaining manipulation performance. Notably, our VT-Gen features a modest model size of 146.74 MiB and achieves an inference speed of 305.90 FPS, ensuring real-time operation and practical deployability.

\begin{table*}[t!]  
    \centering  
    \caption{
    Quantitative results for robotic pushing in simulation. We compare two versions of our ViTacGen framework: one with visual-only inputs and the other with visual \& actual tactile data for reference. These are evaluated against baselines including visual-only, tactile-only, and combined visual-tactile approaches \cite{lin2022tactilegym2}. We conduct 100 trials for each experimental condition. We report cumulative rewards, episode length, distance error, and success rate with a threshold of 2.5 cm. The best results are highlighted in red, while the second-best results are highlighted in blue.
    }  
    \vspace{-1.2em}
    \label{tab:sim_exp}  
    \begin{tabular}{c c c c c c}  
        \toprule  
        Object Type & Method & Rewards (mean$\pm$std) $\uparrow$ & Epi. Len. (mean$\pm$std) $\downarrow$ & Dist. Err. (mm) $\downarrow$ & Succ. Rate (\%) $\uparrow$ \\
        \hline  
        \multirow{5}{*}{Tea Box} & Visual only & -155.73$\pm$63.06 & 335.93$\pm$38.15 & 47.54 & 12.0 \\   
        & Tactile only & -150.10$\pm$62.03 & 337.10$\pm$36.72 & 50.62 & 11.0 \\   
        & Visual \& Tactile &  -147.01$\pm$72.21 & 334.80$\pm$39.34 & 50.02 & 13.0 \\
        & \textbf{Ours} (Visual only) & \color{blue}{-84.35$\pm$79.03} & \color{red}{267.24$\pm$51.90} & \color{blue}{27.81} & \color{blue}{84.0} \\
        & \textbf{Ours} (Visual \& Tactile) & \color{red}{-44.83$\pm$22.64} & \color{blue}{268.57$\pm$30.20} & \color{red}{23.08} & \color{red}{92.0} \\
        \hdashline
        \multirow{5}{*}{Meat Can} & Visual only & -169.27$\pm$80.65 & 330.02$\pm$45.84 & 62.23 & 16.0 \\
        & Tactile only & -149.75$\pm$17.06 & 324.44$\pm$45.39 & 53.77 & 25.0 \\
        & Visual \& Tactile & -125.72$\pm$79.06 & 313.98$\pm$55.18 & 53.75 & 30.0 \\
        & \textbf{Ours} (Visual only) & \color{blue}{-88.19$\pm$79.86} & \color{red}{250.90$\pm$48.50} & \color{red}{29.81} & \color{blue}{81.0} \\
        & \textbf{Ours} (Visual \& Tactile) & \color{red}{-44.49$\pm$20.96} & \color{blue}{251.66$\pm$41.86} & \color{blue}{30.38} & \color{red}{86.0} \\
        \hdashline
        \multirow{5}{*}{Mug} & Visual only & -112.65$\pm$52.30 & 328.78$\pm$43.09 & 54.85 & 20.0 \\   
        & Tactile only & -106.35$\pm$50.17 & 317.51$\pm$49.73 & 47.19 & 31.0 \\   
        & Visual \& Tactile & -106.01$\pm$50.06 & 311.20$\pm$51.95 & 42.83 & 38.0 \\
        & \textbf{Ours} (Visual only) & \color{blue}{-41.53$\pm$19.72} & \color{red}{266.03$\pm$43.91} & \color{blue}{27.39} & \color{red}{86.0} \\
        & \textbf{Ours} (Visual \& Tactile) & \color{red}{-34.92$\pm$13.86} & \color{blue}{270.99$\pm$38.21} & \color{red}{27.07} & \color{blue}{83.0} \\
        \bottomrule  
    \end{tabular}  
    \vspace{-1.2em}
\end{table*}  

\subsection{Simulation Robotic Pushing Results}  
\subsubsection{Quantitative Results}  
As demonstrated in the quantitative results given in Tab. \ref{tab:sim_exp}, simply employing single modality (visual or tactile) fails to learn a good strategy for robotic pushing, as vision only provides a rough estimation of the scene, while tactile sensing only focuses on the interactions between end-effectors and objects, leading to limited perception capability. The Visual \& Tactile baseline shows only marginal improvement because it simply concatenates visual and tactile features without effective cross-modal alignment, which hinders convergence during RL training. Without contrastive learning, the model is unable to establish meaningful relationships between the modalities, resulting in ineffective learning. In contrast, our VT-Con employs contrastive learning and attention-based fusion to properly couple cross-modal representations, facilitating convergence and leading to significant performance improvements across all metrics. Compared to the baseline, ViTacGen with visual and actual tactile data shows an improvement of 64.61\% to 69.50\% on cumulative rewards, 12.92\% to 23.30\% on episode length, and 13.36 mm to 26.94 mm on distance error, and achieves a success rate of 83\% to 94\%. Furthermore, our ViTacGen introduces VT-Gen to eliminate the dependency on tactile modality, instead predicting tactile contact depth images through visual observation. As shown in the table, even with visual modality only as input, ViTacGen maintains good performance across all metrics, significantly outperforming even the baseline with visual and tactile observations, and in some metrics even surpassing its own performance with visual and tactile observations, such as episode length for tea box, meat can, and mug, and the distance error for meat can. It maintains a high success rate of 80\% to 86\%, demonstrating our effectiveness and robustness. 

\begin{figure*} [t!]
	\centering
	\includegraphics[width=0.65 \textwidth]{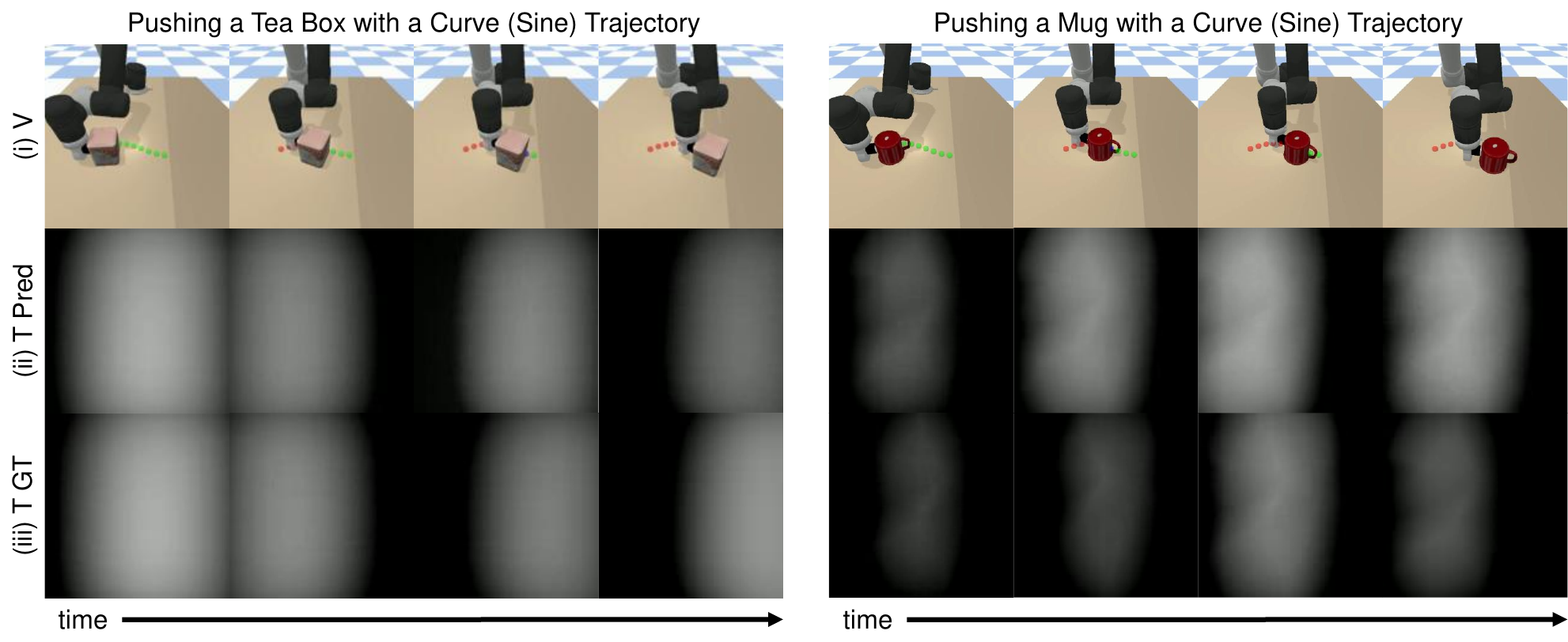}
    \vspace{-1.2em}
	\caption{
    Qualitative results of our ViTacGen in simulation, where we present a tea box and a mug with a curve (sine) trajectories. (i) Visual input with visualized pushing trajectories, where the green points refer to the target trajectory and the red points refer to the successfully executed object trajectory (ii) tactile contact depth prediction generated by VT-Gen (iii) tactile observation. Tactile contact depth maps correspond to frontal end-effector observation. 
    }\label{fig:viz_sim}
    \vspace{-1.2em}
\end{figure*}

\subsubsection{Qualitative Results}
We provide qualitative simulation results of our ViTacGen in Fig. \ref{fig:viz_sim}, where pushing trajectories are visualized in visual images for reference. It can be found in Fig. \ref{fig:viz_sim} (i) that ViTacGen demonstrates robust and stable performance across objects of various shapes. We also present the tactile prediction by VT-Gen and the corresponding tactile observation in Fig. \ref{fig:viz_sim} (ii) and (iii).

\begin{table}[t!]  
    \centering  
    \caption{
    Quantitative results for robotic pushing in real world. Since our ViTacGen in real world setting is deployed on a visual-only robotic system, we compare it against visual-only input baseline in \cite{lin2022tactilegym2}. We conduct 50 trials for each experimental condition and manually measure average distance error and success rate with a threshold of 2.5 cm. 
    }  
    \label{tab:rw_exp}  
    \vspace{-1.2em}
    \begin{tabular}{c c c c}  
        \toprule  
        Object Type & Method & Dist. Err. (cm) $\downarrow$ & Succ. Rate (\%) $\uparrow$ \\
        \hline  
        \multirow{2}{*}{Tea Box} & baseline & 6.5$\pm$1.9 & 14.0 \\   
        & \textbf{Ours} & \textbf{2.6$\pm$0.8} & \textbf{76.0} \\
        \hdashline
        \multirow{2}{*}{Meat Can} & baseline & 8.2$\pm$1.9 & 8.0 \\   
        & \textbf{Ours} & \textbf{1.9$\pm$0.5} & \textbf{82.0} \\
        \hdashline
        \multirow{2}{*}{Mug} & baseline & 7.2$\pm$2.1 & 10.0 \\   
        & \textbf{Ours} & \textbf{1.8$\pm$0.7} & \textbf{86.0} \\
        \bottomrule  
    \end{tabular}  
    \vspace{-1.2em}
\end{table}  

\begin{figure*} [t!]
	\centering
    \vspace{0.4em}
	\includegraphics[width=0.65 \textwidth]{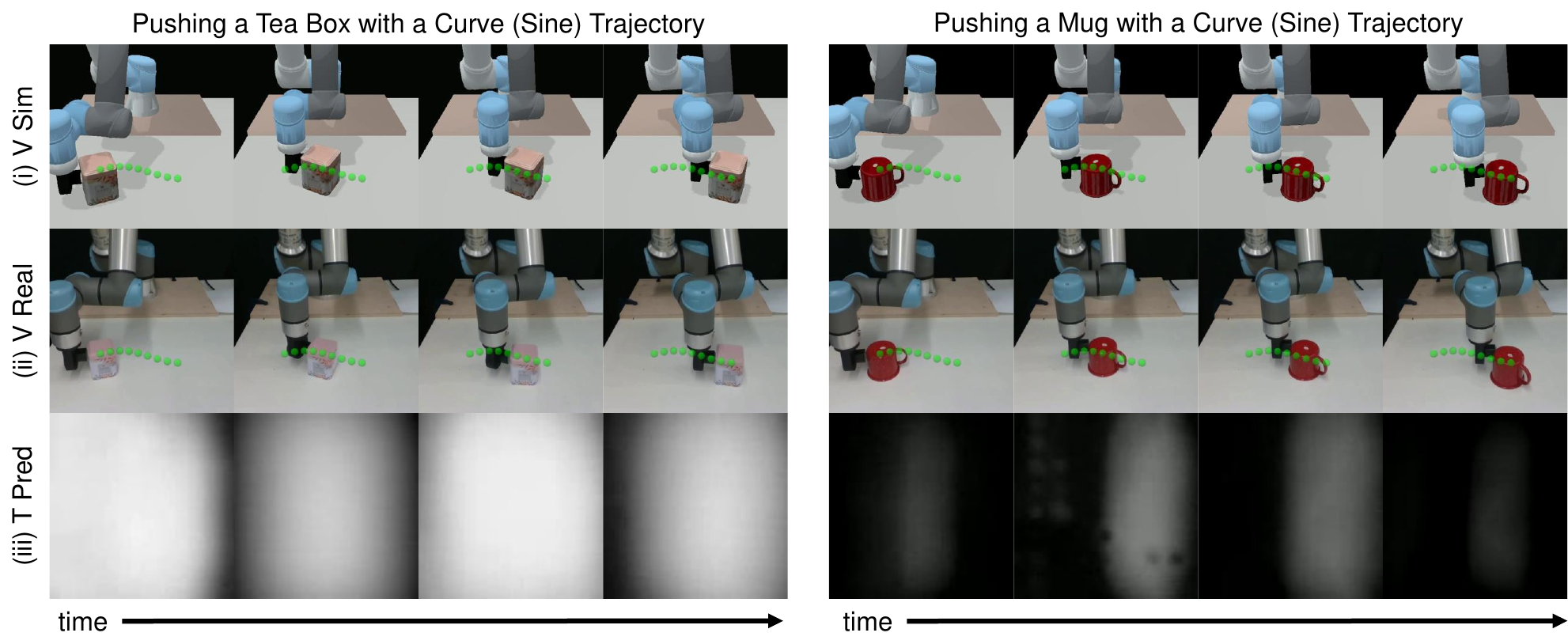}
    \vspace{-1.2em}
	\caption{
    Demonstration of ViTacGen in real world scenarios, where we present a tea box and a mug with a curve (sine) trajectories. The trajectories are visualized in simulation and projected onto both simulated and real world environments to demonstrate the system's performance. (i) Simulation visual sequences for reference (ii) real world visual sequences (iii) tactile contact depth images generated from real world visual input, corresponding to frontal end-effector observation. More qualitative results are available in our supplementary material. 
    }\label{fig:viz_real}
    \vspace{-1.2em}
\end{figure*}

\subsection{Real World Robotic Pushing Demonstration}
The performance of ViTacGen in real world scenarios is evaluated both quantitatively and qualitatively, as presented in Tab. \ref{tab:rw_exp} and Fig. \ref{fig:viz_real}, respectively. We conduct 50 trials for each experimental condition and manually measure average distance error and success rate. In Fig. \ref{fig:viz_real}, we visualize the trajectories in simulation and their projections onto both simulated and real world environments to demonstrate the system's effectiveness. The experimental results indicate that ViTacGen achieves robust zero-shot transfer to real world scenarios, primarily due to VT-Gen's capability in capturing subtle contact forces and dynamics between the end-effector and the object. Although the tactile predictions for real world scenarios exhibit some noise artifacts (Fig.\ref{fig:viz_real} (iii)), attributable to the sim-to-real discrepancy in visual modalities, as evident in Fig.\ref{fig:viz_real} (i) and (ii), these perturbations do not significantly impact the RL algorithm's execution. Notably, the distance errors in real world experiments are consistently lower than their simulation counterparts, which can be attributed to our implementation of a threshold-based termination criterion in the simulation environment. We also provide real world robotic pushing results on unseen objects in Tab. \ref{tab:unseen}, where our ViTacGen demonstrates good generalizability and robustness in handling unseen objects with diverse shapes. On average, our method achieves 75.2\% success rate with 3.8 cm distance error across these diverse unseen objects, with the best performance on the apple, which is the smallest object among all the objects, which is easiest to control during pushing, and the worst on the soup can, likely due to its high center of gravity and tendency to spin during pushing motions. More qualitative results are available in our demo video. 

\begin{table}[t!]  
    \centering  
    \caption{
    Quantitative results for pushing unseen objects in real world. We conduct 50 trials for each experimental condition and manually measure average distance error and success rate with a threshold of 4.0 cm. 
    }  
    \label{tab:unseen}  
    \vspace{-1.2em}
    \begin{tabular}{c c c c}  
        \toprule  
        Object Type & Dist. Err. (cm) $\downarrow$ & Succ. Rate (\%) $\uparrow$ \\
        \hline  
        Apple & 3.3$\pm$1.1 & 82.0 \\   
        Coffee Box & 4.1$\pm$2.0 & 72.0 \\
        Ceramic Cup & 3.9$\pm$1.8 & 74.0 \\
        Olive Jar & 3.7$\pm$1.7 & 78.0 \\
        Soup Can & 4.1$\pm$1.7 & 70.0 \\
        \bottomrule  
    \end{tabular}  
    \vspace{-1.2em}
\end{table}  

\begin{table*}[t!]  
    \centering  
    \vspace{0.4em}
    \caption{
    Ablation study on the fusion module. We conduct 100 trials for each experimental condition. We report cumulative rewards, episode length, distance error, and success rate with a threshold of 4.0 cm. 
    }  
    \vspace{-1.2em}
    \label{tab:fusion}
    \begin{tabular}{c c c c c}  
        \toprule  
        Fusion Module & Rewards (mean$\pm$std) $\uparrow$ & Epi. Len. (mean$\pm$std) $\downarrow$ & Dist. Err. (mm) $\downarrow$ & Succ. Rate (\%) $\uparrow$ \\
        \hline  
        Addition & -70.00$\pm$36.47 & 265.01$\pm$47.70 & 39.93 & 77.0 \\
        Concatenation & -67.84$\pm$59.26 & 251.81$\pm$35.21 & 36.37 & 90.0 \\
        Attention \cite{dosovitskiy2020vit} & \textbf{-66.92$\pm$55.28} & \textbf{247.82$\pm$27.81} & \textbf{35.89} & \textbf{95.0} \\
        \bottomrule  
    \end{tabular}  
    \vspace{-1.2em}
\end{table*}  

\begin{table*}[t!]  
    \centering  
    \caption{
    Ablation study on the contrastive learning module. We conduct 100 trials for each experimental condition. We report cumulative rewards, episode length, distance error, and success rate with a threshold of 4.0 cm. 
    }  
    \vspace{-1.2em}
    \label{tab:cl}
    \begin{tabular}{c c c c c}  
        \toprule  
        Fusion Module & Rewards (mean$\pm$std) $\uparrow$ & Epi. Len. (mean$\pm$std) $\downarrow$ & Dist. Err. (mm) $\downarrow$ & Succ. Rate (\%) $\uparrow$ \\
        \hline  
        SimCLR \cite{chen2020simclr} & -70.03$\pm$49.30 & 251.97$\pm$36.26 & 37.58 & 89.0 \\
        MoCo \cite{he2020moco} & \textbf{-66.92$\pm$55.28} & \textbf{247.82$\pm$27.81} & \textbf{35.89} & \textbf{95.0} \\
        \bottomrule  
    \end{tabular}  
    \vspace{-1.8em}
\end{table*}  

\vspace{-0.4em}
\subsection{Ablation Study}
We conduct ablation studies on fusion module and contrastive learning module. The results are shown in Tabs. \ref{tab:fusion} and \ref{tab:cl}. For fusion module, we compare addition, concatenation, and attention \cite{dosovitskiy2020vit}, demonstrating that attention-based fusion achieves the best performance. For the contrastive learning module, we compare SimCLR \cite{chen2020simclr} and MoCo \cite{he2020moco}, showing that MoCo outperforms SimCLR across all metrics. 

\section{Conclusion}
This paper presents ViTacGen, a novel robot manipulation framework designed for visual robotic pushing with vision-to-touch generation in reinforcement learning to eliminate the reliance on high-resolution real tactile sensors, inspired by human's remarkable ability of predicting tactile states from vision to optimize manipulation. Specifically, we introduce an encoder-decoder vision-to-touch generation network that generates contact depth images, a standardized tactile representation, directly from visual image sequence, followed by a reinforcement learning policy that fuses visual-tactile data with contrastive learning based on visual and generated tactile observations. ViTacGen enables zero-shot deployment on visual-only robotic systems. We conduct extensive evaluation in both simulated and real world environments, demonstrating the effectiveness of our proposed methods. 

While our ViTacGen demonstrates effectiveness, it contains certain limitations. Our method employs contact depth as a high-level tactile representation to provide a perception for contact geometry rather than generating precise tactile images, which reduces the sim-to-real gap but limits our ability to perceive detailed physical properties such as local force distributions and fine-grained surface geometry. Future work could focus on explicitly predicting these physical properties to enhance policy performance. Additional promising research direction involves exploring methods to infer shear forces \cite{lin2020invariant} applied on tactile sensors directly from visual observations of object motions, which would further bridge the gap between visual and tactile modalities. 

\section*{Acknowledgment}
The authors would like to thank Yupeng Wang for his help in hardware support for the experiments.

{
\tiny
\bibliographystyle{IEEEtran}
\bibliography{main}
}

\end{document}